\documentclass{article}

\usepackage[nonatbib,final]{neurips_2019}

\usepackage[utf8]{inputenc}
\usepackage[T1]{fontenc}
\usepackage{hyperref}
\usepackage{url}
\usepackage{booktabs}
\usepackage{CJKutf8}

\usepackage{amsfonts}
\usepackage{nicefrac}
\usepackage{microtype}
\usepackage{graphicx}
\usepackage{xcolor}
\usepackage{lipsum}

\title{
    Investigating Effect of Dialogue History in Multilingual Task Oriented Dialogue Systems
  \vspace{1em}
}

\author{
  Michael Sun \\
  Department of Computer Science \\
  Stanford University \\
  \texttt{msun415@cs.stanford.edu} \\
  \And
  Kaili Huang \\
  Department of Computer Science \\
  Stanford University \\
  \texttt{kaili@cs.stanford.edu} \\
  \And
  Mehrad Moradshahi \\
  Department of Computer Science \\
  Stanford University \\
  \texttt{mehrad@stanford.edu}
}

\begin{document}

\maketitle

\begin{abstract}
While the English virtual assistants have achieved exciting performance with an
enormous amount of training resources, the needs of non-English-speakers have not been satisfied well. Up to Dec 2021, Alexa, one of
the most popular smart speakers around the world, is able to support 9 different
languages\cite{alexa}, while there are thousands of languages in the world, 91 of which
are spoken by more than 10 million people according to statistics published in 2019\cite{wiki-languages}. However, training a virtual assistant in other languages than English is often more difficult, especially for those low-resource languages. The lack of high-quality training data restricts the
performance of models, resulting in poor user satisfaction. Therefore, we devise an efficient and effective training solution
for multilingual task-orientated dialogue systems, using the same dataset generation pipeline and end-to-end dialogue system architecture as the SOTA paper BiToD\cite{bitod}, which adopted some key design choices for a minimalistic natural language design where formal dialogue states are used in place of natural language inputs. This reduces the room for error brought by weaker natural language models, and ensures the model can correctly extract the essential slot values needed to perform dialogue state tracking (DST). Our goal is to reduce the amount of natural language encoded at each turn, and the key parameter we investigate is the number of turns (H) to feed as history to model. We first explore the turning point where increasing H begins to yield limiting returns on the overall performance. Then we examine whether the examples a model with small H gets wrong can be categorized in a way for the model to do few-shot finetuning on. Lastly, will explore the limitations of this approach, and whether there is a certain type of examples that this approach will not be able to resolve. 
 
\end{abstract}

 


\section{Motivation and Problem}


While the English virtual assistants have achieved exciting performance with an enormous amount of training resources, non-English-speakers' needs to interact with virtual assistants have not been satisfied well. Up to Dec 2021, Alexa, one of the most popular smart speakers around the world, is able to support 9 different languages \cite{alexa}, while there are thousands of languages in the world, 91 of which are spoken by more than 10 million people according to statistics published in 2019 \cite{wiki-languages}. Therefore, the importance of multilingual virtual assistants can never be overemphasized. Moreover, while people are interacting with virtual assistants, it often occurs that they want to refer to certain entities (e.g., Chinese restaurants, Korean songs, etc.) in their original languages.

However, training a virtual assistant in other languages than English is often more difficult, especially for those low-resource languages. The lack of high-quality training data often restricts the performance of models, resulting in poor user satisfaction. Therefore, it is very important to devise efficient and effective training methods for multilingual virtual assistants.

\section{Related Work}
\subsection{End-to-end Dialogue Systems}








To build a conversational virtual assistant, a straightforward practice is to integrate multiple modules (e.g. Language Understanding, Dialogue Policy, etc.) into a dialogue system, where different modules are trained separately to deal with different parts of the dialogue. However, this type of modulated task-oriented dialogue systems are prone to accumulated errors. Since the first end-to-end system, where all components of the system are trained at the same time, was developed\cite{Li2017EndtoEndTN}, a lot of progress has been made to get better overall performance and while keeping the system compact. The state-of-the-art method we build on uses a single seq2seq language model to simultaneously perform each module of the virtual assistant.

\subsection{Multilingual Dialogues Via Translation}



\begin{CJK*}{UTF8}{gbsn}
Given a large amount of training data for English virtual assistants, we can directly use Neural Machine Translation (NMT) to generate a parallel dataset in another language. However, challenges include entities often being mistranslated/transliterated/dropped. Multiple methods can be adopted to improve the quality of translation. For example, in order not to mess up the entities' translation (e.g., a restaurant called ``Lakeside Dining'', a dining hall at Stanford, should not be translated to ``湖边的餐厅'', which means a restaurant near the lake), we can first assign each entity a standard translation by using Google Translate followed by human correction, and then translate the complete dialogue \cite{dstc9}. However, this does not assure accurate translation. Moreover, annotation and translation errors can compound over multiple turns. Thus, translation with alignment is often not enough to obtain a high-quality multilingual dataset.
\end{CJK*}

\subsection{BiToD}









One of the state-of-the-art method is proposed in the BiToD paper \cite{bitod}. BiToD uses a Seq-to-Seq model such as mBART\cite{mbart} that can be trained on multiple tasks simultaneously. The model is trained on DST, Agent Policy and Agent Response tasks when given input data with according prompts. In the end-to-end setting, at each step, the prediction of the previous step, together with all the dialogue history (including user utterances and agent acts) are given as inputs to the model to make prediction for the current step (e.g., updated dialogue state, whether to make an API call, knowledge base query, etc). In the turn-by-turn evaluation setting, at each step, the gold input is given as input instead, and will be used to validate and improve the model.

\begin{figure}[h]

\centering

\includegraphics[width = 8cm ]{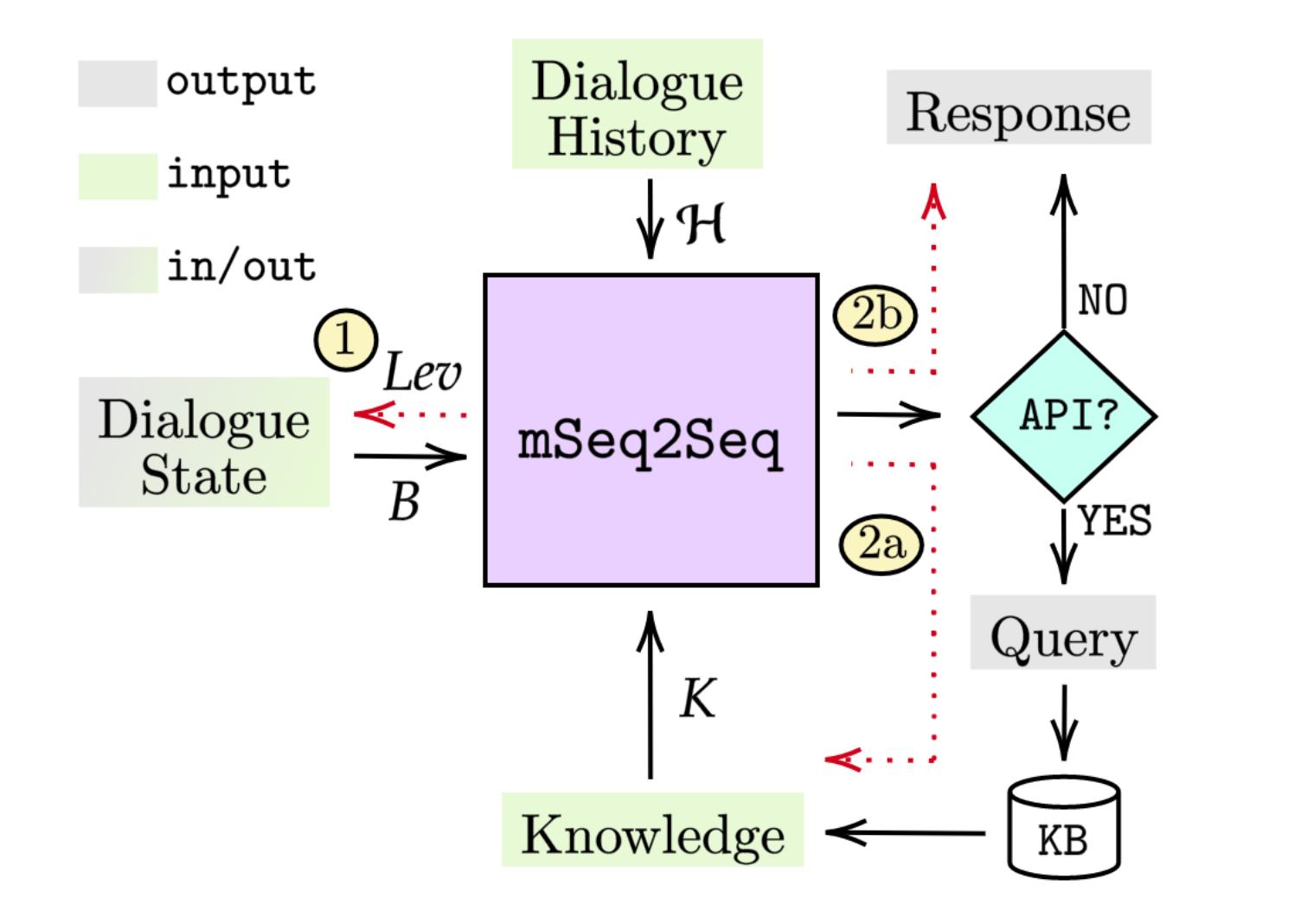}

\end{figure}

\subsection{Dataset}
 Our project aims to understand the challenges and devise solutions for multilingual task-orientated dialogue systems. We work with the dataset from the BiToD paper, where dialogues are generated from a dialogue simulator that interacts with APIs on a bilingual knowledge base, and then paraphrased via crowdsourcing. The entities map closely between English and Chinese, but the simulated dialogues are not constructed in parallel. This pipeline already avoids the pains associated with translation, but still have to grapple with a few of the aforementioned difficulties of multilingual systems: less crowdsource workers for paraphrasing, lower annotation quality and weaker natural language models. Due to the last factor, the option for neural generation module is also impractical. Therefore, translated templates are used for natural language generation instead. 

\section{Approach/Idea}
We believe BiToD's work has a key design choices which encode the philosophy we take to this project: a minimalistic natural language design where formal dialogue states are used in place of natural language inputs. This reduces the room for error brought by weaker natural language models, and ensures the model can correctly extract the essential slot values needed to perform its task. 
Due to BiToD's parallel design between English and Chinese, the authors showed a bilingual model, trained on both English and Chinese data performed stronger than either of the two trained independently. That indicates the advantage of using formal dialogue states to model's transfer learning abilities.

However, together with the formal dialogue states, this design also accumulates all historical natural language utterances as part of the model's inputs.
We wanted to further optimize the model by constraining the place where natural language is fed as input to the system: dialogue history.

\subsection{Solution to Challenge}
Our goal is to reduce the amount of natural language encoded at each turn, and the key parameter we investigate is the number of turns ($H$) to feed as history to model. We first explore the turning point where increasing $H$ begins to yield limiting returns on the overall performance. Then we examine whether the examples a model with small $H$ gets wrong can be categorized in a way for the model to do few-shot finetuning on. Lastly, will explore the limitations of this approach, and whether there's a certain type of examples that this approach won't be able to resolve.

The agent responses have been replaced with formal dialogue acts, so the only natural language encoded from the history will be the user utterances. The formal dialogue state is fed as input to each step in every turn: ``dst'' (to obtain the state update $lev$), ``api'' (the agent policy, outputting whether to call api or go to response), and ``response'' (dialogue act). 

We use a separate NLG module, decoupled from the end-to-end model to eliminate error accumulation. And we use standard translated templates for agent response generation.

\subsection{Issues to Address}
\label{4.2}
This is an example of long-range dependency. When the agent fails to find a place that satisfies all the user's requirements, it asks to change the \textbf{location} requirement. The user accepts the suggestion and adds another requirement about the price. If we only feed in one historical input (i.e. the user input in this turn), the agent can update ``PriceRange'' but cannot update ``Location'' correctly because it cannot refer to the suggestion it makes in the previous turn. This example gives an explanation why more history is necessary in some cases.

\includegraphics[scale=0.42]{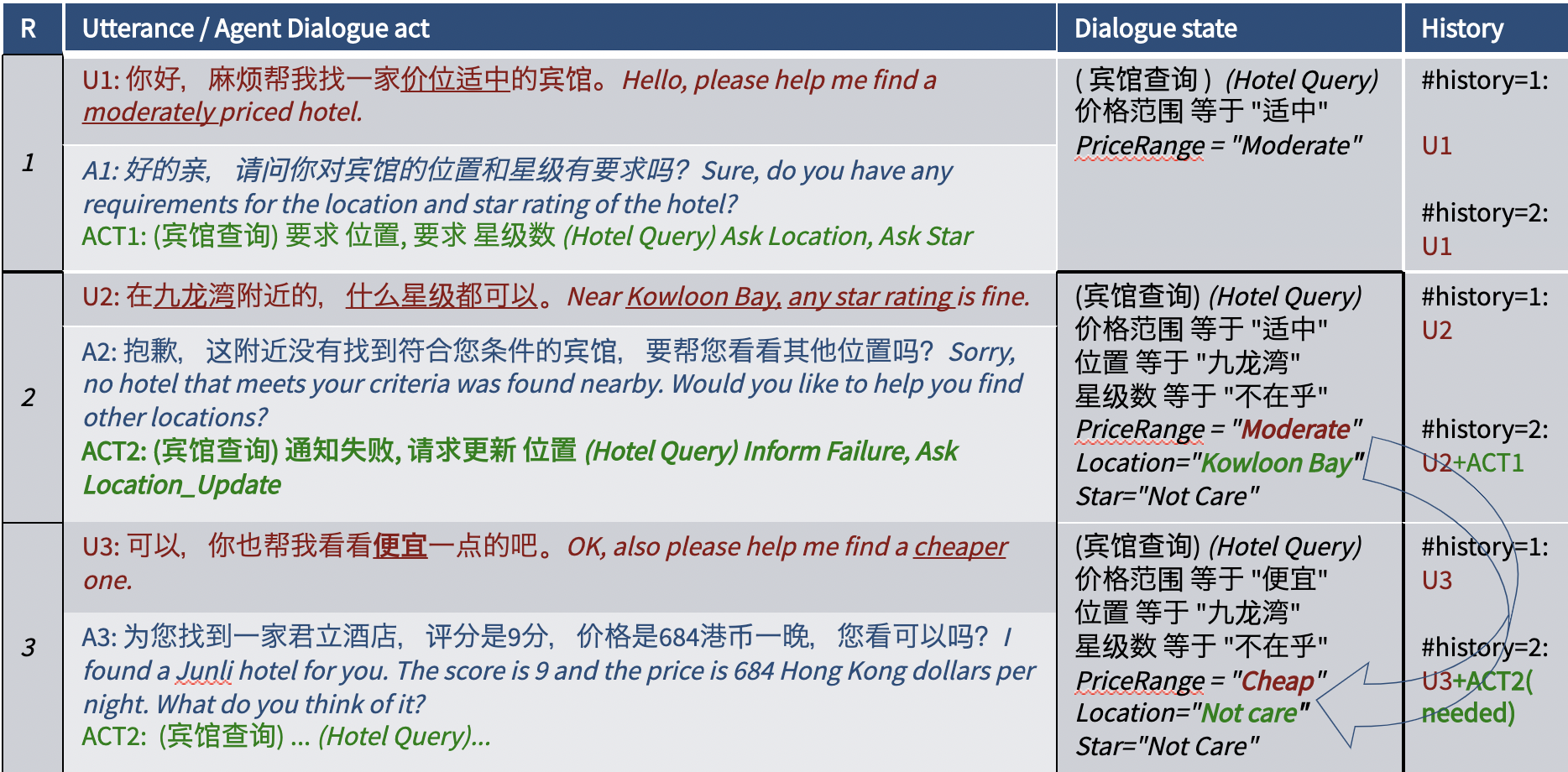}

\section{What We Have Done}
In the data preprocessing step, we generate different datasets by adjusting the parameter $H$, which refers to the maximum number of turns that can be encoded as part of the model's input. Each turn can be a user utterance or agent dialogue act. For example, $H=1$ means only the previous turn's user utterance is encoded (together with the current agent act, which is always encoded for all model), and $H=2$ means additionally encoding the previous turn's agent act, and so forth. Since we want to investigate the feasibility of reducing the amount of natural language encoded, we run extensive experiments for different values of $H$ on both the whole dataset and a few-shot dataset which compiles examples including long range dependencies like the one in 4.2.

\subsection{Initial Experiment Procedure}

We preprocess the data for $H=1,2,3,4$ and put them in separate directories.

We used genienlp's train module to run our initial experiments. We train an end-to-end Chinese to Chinese model using Facebook's mbart large 50 as our pretrained seq2seq model. All models are trained for 50000 iterations. We use a batch size of 800, validation batch size of 4000. We use ADAM optimizer for all experiments and use transformers learning rate scheduler with a max learning rate of 0.01. 

\subsection{Few-shot Dataset Curation}

\begin{CJK*}{UTF8}{gbsn}
We curated a few-shot dataset from the $H=1$ directory involving examples with long-range dependencies that require a $H\ge 2$ model to disambiguate. To do so, we look at one representative dialogue pattern: 
examples that involve the agent making concrete recommendations or proposing alternative slots. This is almost always the case after the agent outputs an ``通知失败 , 请求更新'' (i.e. ``inform failure, request update'') dialogue act. In both cases, the agent needs to remember what it said (i.e. its previous dialogue act) in order to have context for the user's response. Here is an example.
\end{CJK*}

\begin{itemize}
    \item User: Find me 5-star hotel in LA under \$100/night.
    \item Agent: \textit{Sorry}, no hotels satisfy your requirements. (1) \textit{Can I recommend some 3-star hotels instead?} (2)
    \item User: I'm cool with \textit{that} (3)
    \item Agent: …? (history: [User: I'm cool with that])
\end{itemize}

\begin{CJK*}{UTF8}{gbsn}
Note that (2: makes recommendation or suggests alternative) and (3: responds in affirmation, but does not repeat the specific slot and value) define this pattern of examples. If the user did repeat the agent's recommendation (e.g. ``I'm cool with 3-star hotels''), then the agent did not need to refer to its historical dialogue act. (1: inform failure) doesn't necessarily need to occur, but most often does. For that practical reason, we sourced our few-shot dataset by matching for all dialogue\_ids for which the ``通知失败 , 请求更新'' dialogue act appeared, since for all cases we found, the ``通知失败， 请求更新'' dialogue act is followed by (2). Then, we manually narrowed the set down to only where (3) holds. There are many cases where the user does mention the agent's recommendation, but not clearly enough to disambiguate. We include these examples as well, since they may help make our $H=1$ model more robust.
\end{CJK*}

Lastly, we split the few-shot dataset into train and validation splits. In total, our few-shot dataset has 73 and 6 dialogues, 987 and 63 turns in train and validation respectively.

\subsection{Few-shot Finetuning Procedure}

Our fine-tuning procedure is very similar to the training procedure. We fix the same parameters for a fair comparison, but validate and save the results every 2 (instead of 1000) steps. All models are trained for 100 iterations.

\subsection{Case Analysis of Examples}




To analyze the difference between $H=2,3,4$ models, we investigate the set of dialogue turns where each model gets wrong (error set) and focus on the relationship of their error sets. Let $S_2, S_3, S_4$ denote the error sets of $H=2,3,4$ models, respectively. The size of the error sets and their intersections are shown in table \ref{tab:bad-set}.

\begin{table}[h]
    \centering
    \begin{tabular}{|l|l|}
    \hline
    \textbf{Set} & \textbf{Size of Set} \\ \hline
       $ S_2$  &  1335 \\ \hline
        $ S_3$   & 1304 \\ \hline
         $ S_4$ & 1189 \\ \hline
         $ S_2 \cap S_3$  &  1001 \\ \hline
         $ S_3 \cap S_4$  &  954 \\ \hline
         $ S_2 \cap S_4$  &  714 \\ \hline
    \end{tabular}
    \caption{Sets of Turns Where Models Get Wrong}
    \label{tab:bad-set}
\end{table}

We evaluate $H=3$ model on the turns that $S_2$ gets wrong. The Exact Match (EM) score is $25\%$. This can also be calculated from figures in the table (${\vert S_2-S_2 \cap S_3 \vert}\div {\vert S_2\vert}=25\%$). Inversely, we also evaluate $H=2$ model on $S_3$, and the EM score is $23\%$, which is close to $25\%$. This indicates that although $H=3$ model has a better performance on some samples than $H=2$ model, it also does worse than $H=2$ model on some other samples. And the ``increase'' on performance (indicated by $25\%$) is very close to the ``decrese'' in performance (indicated by $23\%$). Then we reach the preliminary conclusion that from $H=2$ model to $H=3$ model, we do not obtain a real improvement. The things are different for $H=4$ model, however. The ``increase'' and ``decrease'' from $H=3$ model to $H=4$ model are $27\%$ and $18\%$, respectively. And ``increase'' and ``decrease'' from $H=2$ model to $H=4$ model are $47\%$ and $40\%$, respectively. 

\subsubsection{Dialogues That Require 3 Turns}
By digging deeper into the dialogue turns where $H=2$ model makes mistakes and $H=3$ models does not, we find that there is not a mistake pattern related to the ``long-range dependency'' mentioned in section \ref{4.2}. Take $H=2$ model as an example. In the turns that $H=2$ model gets wrong and $H=3$ model gets correct ($S_2-S_2\cap S_3$), $30\%$ are wrong predictions on whether to make an API call or not. $57\%$ are mistakes made in the response step and the mistake patterns are mainly related to different format of date (e.g., Monday v.s. Dec 6th), asking for different parameters to make a certain query (e.g. for restaurant reservation, number of people v.s. reservation time), etc. The other $23\%$ mistakes are made in the DST step, where a frequent mistake pattern is when the user asks ``Is there a hotel near that restaurant?'', the model cannot interpret ``near that restaurant'' to a correct location requirement, despite the dialogue state containing the restaurant reservation from earlier in the dialogue. This pattern is thus not a long-dependency issue that requires $H=3$.

\subsubsection{Dialogues That Require 4 Turns}
For dialogue turns that $H=2$ model makes mistake and $H=4$ model does not, we find an interesting pattern. A simplified real example is shown below.

\begin{itemize}
    \item User: Can you find me a restaurant near with more than 5 stars?
    \item Agent: Yes, I found 22 for you. I recommend Emporio Antico. 
    \item User: Can you recommend another one? I want a cheaper one.
    \item Agent: Sure, I recommend DiVino Patio.
    \item User: What's the address?
    \item Agent: The address is ...
    \item Agent: OK, please reserve \textbf{it} for me. (``it''=DiVino Patio)
\end{itemize}

In this example, after the agent gives some suggestion (DiVino Patio), the user further asks information about the suggestion. And after that, when the user wants to make a reservation, the agent does not know which restaurant to reserve unless the model is fed at least 4 historical inputs ($H=4$).

After going through the diff between $H=2$ model's mistakes and $H=4$ model's, we find a few such examples (more than 50 turns, according to rough estimation). To conclude, after analyzing the models' error types, we do not observe examples that require $H=3$, but we observe a few examples that require $H\ge 4$. This should be a byproduct of the dialogue simulation pipeline of the BiToD dataset.

The scenarios shown in the above example could hypothetically get more complex and the range of the dependency arbitrarily long. If the user asks a lot about the suggestion and then asks the agent to make a reservation, the agent will need to refer to a large amount of historical turns. However, feeding arbitrarily many historical inputs does not seem an optimal way of solving this. A better solution is to change the format of dialogue state, so that previous agent acts should also be properly kept track of, and we will continue this discussion in section 9. 

We will fine-tune on the much more common $H=2$ dependency examples to improve our $H=1$ model, as examples that have longer dependencies are too few to learn from. 

\section{Results}

\begin{table}[h]
\centering
\begin{tabular}{|l|l|l|}
\hline

\textbf{Max History (H)} & \textbf{Exact Match (EM)} & \textbf{Cased BLEU} \\ \hline
H = 1           & 78.6             & 93.8       \\ \hline
H = 2           & 82.6             & 95.2       \\ \hline
H = 3           & 83.0             & 95.0       \\ \hline
H = 4           & 84.5             & 95.5       \\ \hline
\end{tabular}
\caption{\label{tab:table1}Whole Dataset}
\end{table}

\begin{table}[h]
\centering
\begin{tabular}{|l|l|l|}
\hline
\textbf{Max History (H)} & \textbf{Exact Match (EM)} & \textbf{Cased BLEU} \\ \hline
H = 1             & 40.7             & 57.4         \\\hline
H = 1 w/ few-shot & 62.4 (+21.7)     & 73.9 (+16.5) \\\hline
H = 2             & 79.4             & 94.7         \\\hline
H = 3             & 83.0             & 95.0        \\\hline
\end{tabular}
\caption{\label{tab:table2}Few-shot Dataset}
\end{table}

\begin{table}[h!]
\centering
\begin{tabular}{|l|l|l|}
\hline
\textbf{Max History (H)} & \textbf{Exact Match (EM)} & \textbf{Cased BLEU} \\ \hline
H = 1                    & 78.6                      & 93.8                \\ \hline
H = 1 w/ few-shot        & 78.6                      & 93.7                \\ \hline
H = 2                    & 82.6                      & 95.2                \\ \hline
H = 3                    & 83.0                      & 95.0                \\ \hline
H = 4           & 84.5             & 95.5       \\ \hline

\end{tabular}
\caption{\label{tab:table3}Whole Dataset}
\end{table}

For these experiments, we evaluate in the turn-by-turn setting, where the gold input is fed in at every turn.


\section{Findings}
Consistent with our initial motivations for reducing the amount of natural language encoded, we see diminishing returns of storing more history. On the whole dataset, the $H=3$ model only gets a 0.4 increase in EM, and a 0.2 decrease in Cased BLEU, putting in question the merit for extending $H$ beyond $2$. We found most examples don't even require 2 turns of history. Despite being extra lenient with some examples for the purposes of making our $H=1$ more robust, our few-shot data amounts to just under ~1\% of the whole dataset.

Most amazingly, we discovered that for the few examples that do reference many turns ago, few-shot finetuning a low (max history) model can easily match a high (max history) model without sacrificing overall accuracy!

We can propose an optimal trade-off point for $H$(=1) to store which would balance between our goals of reducing natural language footprint with maintaining strong performance after fine-tuning.

\section{What did you learn?}
We learned that more natural language doesn't always help. Less natural language means less room for error, more transferable formal representation, and more flexibility for generalization. Using the minimal input needed helps reduce both training footprint and allow easy generalization via fine-tuning to harder dialogues. 

We learned that the formal representation of the dialogue state is especially relevant in the multilingual setting, as it is the common language which all languages share, regardless if they're a low-resource language or a globally spoken one. 

We also stumbled across a benefit of data created from simulation, since we were quickly able to reverse-engineer the examples to find a canonical form of examples that require >= 2 turns of history, namely to the dialogue act the agent outputted. Had the dialogues been naturally generated, that would not have been possible. We found an underrated advantage of using synthetic data generation is that it can help make debugging multilingual agents easy even for non-native speakers in the future.

\section{Future work}
A future extension of our current fine-tuning methodology would be to find harder few-shot data with longer dependencies. As we have reached the limit of BiToD's dataset existing simulation pipeline, it is time to look for extensions of it that can sample user goals which are contingent on one another and is often separated by at least $H=4$ turns of history.

At the same time, we fully embrace alternative ways to encode history. As mentioned before, we can properly encode agent acts into the dialogue state to directly track the agent's recommendation/suggestion and resolve the long-range dependency examples in the BiToD dataset which require $H\ge 4$ turns of history.

Alternatively, we can relax the current slot-value-relation representation and keep miscellaneous text spans that would seem relevant for a later dialogue state. For example, suppose the user mentions a curious fact about their preferences early in the conversation e.g. ``I really love burgers, but I'm on a diet'' then after being dissatisfied with the agent's healthier recommendations later says ``well I only live once, so get me some fast food place now!'' The agent would be expected to use the earlier mention to find burger restaurants. This could involve a dynamic input/output/forget gates like in LSTM or other attention-based networks, which would train the agent to dynamically update parts of its own dialogue state beyond accumulated slot-value-relation's. These forms of alternate dialogue state representation require more thought but we'll be open to learning more should we be given the opportunity to continue this line of work in the future!

\section{Contributions}

This project saw equal contributions between Michael and Kaili. Both worked very hard to make the most out of it and found it a very rewarding experience! Mehrad was a helpful mentor who introduced us to relevant resources and unblocked us when we got stuck. He also proofread this report.
\newpage
\bibliographystyle{unsrt}

\end{document}